\documentclass{article}
\usepackage{spconf,amsmath,graphicx}
\usepackage{longtable}
\usepackage{subfigure}
\usepackage{paralist}
\usepackage{booktabs}
\usepackage{enumitem}
\usepackage{stackengine}
\usepackage{alltt}
\usepackage{esvect}
\usepackage{listings}
\usepackage[usenames,dvipsnames]{xcolor}
\usepackage{numprint}
\usepackage[hyphens]{url}
\usepackage{pdflscape}
\usepackage{mathtools, bm, amsfonts, amssymb}
\usepackage{paralist}
\usepackage{caption}
\usepackage{balance}

\newcommand{\etal}{\textit{et al.}}

\title{Combining multimodal information for Metal Artefact Reduction: an unsupervised deep learning framework}
\name{\parbox{\linewidth}{\centering Marta B.M. Ranzini$^{1,2}$ \qquad Irme Groothuis$^{2}$ \qquad Kerstin Kl{\"a}ser$^{1,2}$ \qquad M. Jorge Cardoso$^{2}$ \qquad \textit{Johann Henckel$^{3}$ \qquad S\'ebastien Ourselin$^{2}$ \qquad
Alister Hart$^{3}$ \qquad Marc Modat$^{2}$}}}
\address{$^{1}$ Medical Physics and Biomedical Engineering Department,
University College London, UK \\
$^{2}$ School of Biomedical Engineering \& Imaging Sciences, King's College London, UK \\
$^{3}$ Royal National Orthopaedic Hospital NHS Trust, UK}

\begin{document}
%
\maketitle
\begin{abstract}
Metal artefact reduction (MAR) techniques aim at removing metal-induced noise from clinical images. In Computed Tomography (CT), supervised deep learning approaches have been shown effective but limited in generalisability, as they mostly rely on synthetic data. In Magnetic Resonance Imaging (MRI) instead, no method has yet been introduced to correct the susceptibility artefact, still present even in MAR-specific acquisitions.
In this work, we hypothesise that a multimodal approach to MAR would improve both CT and MRI. Given their different artefact appearance, their complementary information can compensate for the corrupted signal in either modality.
We thus propose an unsupervised deep learning method for multimodal MAR. We introduce the use of Locally Normalised Cross Correlation as a loss term to encourage the fusion of multimodal information. 
Experiments show that our approach favours a smoother correction in the CT, while promoting signal recovery in the MRI.
\end{abstract}
\begin{keywords}
Metal Artefact Reduction, CT, MR, Deep Learning, Unsupervised Learning
\end{keywords}
%


\section{Introduction}
\label{sec:intro}
Metallic implants are one of the main causes for image quality degradation in medical imaging. 
In Computed Tomography (CT), the metal's higher attenuation coefficient causes signal corruption, resulting in bright and dark streaks that irradiate from the metal source throughout the reconstructed image. 
In Magnetic Resonance Imaging (MRI), metal objects induce local magnetic field inhomogeneities that cause intensity and geometrical distortions in the reconstructed image. These susceptibility artefacts typically appear as blackened areas close to the implant, partially shadowing the neighbouring structures. \\
In patients with hip replacement, the size of the metallic prosthesis makes the artefacts even more severe and extended \cite{Gjesteby2016}, hampering the diagnostic interpretation of the images in the most clinically relevant areas, i.e. close to the implant. As a result, the introduction of successful metal artefact reduction (MAR) techniques in hip replacement imaging is of great importance and an active field of research. \\
Numerous approaches have been proposed in the literature for MAR in CT \cite{Gjesteby2016}. Traditional physics-based or iterative reconstruction methods are now being challenged by novel deep neural network approaches, which are data-driven and less dependent on physical model assumptions. However, most methods \cite{Wang2018,Gjesteby2018} are trained in a supervised fashion, relying on either pre- and post-operative paired data (not always available) or simulations (not realistic enough).
A solution to the supervised training setting was recently proposed by Liao \etal~\cite{Liao2019}. They introduced an unsupervised adversarial training scheme to disentangle the artefact from the anatomy appearance in CT images, showing state-of-the-art performance on both synthetic and real data.
In MRI research, efforts have focused mostly on image acquisition improvements: Tailored MR sequences such as MARS \cite{Olsen2000} or SEMAC \cite{Lu2009} have proven effective in reducing the extension of the shadowing, but cannot completely eliminate it, making the clear visualisation of the implant in MRI impossible. \\
In this work we introduce a novel unsupervised deep learning MAR method for \textit{jointly} correcting same-subject CT and MR hip images. Our Multimodal Artefact Disentanglement Network (MADN) extends the approach proposed by Liao \etal~\cite{Liao2019} by introducing a similarity loss that induces the network to learn shared information between CT and MRI content. As a result, the CT correction takes advantage of the sharper contrast of MRI throughout the field of view, while the MRI correction is helped by implant localisation information from the CT. As the appearance of the artefact is different in CT and MRI, we believe that making use of their contextual complementary information would help better correct for the artefact in both modalities. 
We demonstrate the effect of our approach using intensity distribution analysis in the CT and on a segmentation propagation task in MRI, showing how multimodal information improves the correction compared to single-modality approaches.

\section{Method}
\label{sec:method}
\subsection{Artefact Disentanglement Network} 
\begin{figure*}
\begin{center}
  \includegraphics[height=5cm, width=15.5cm]{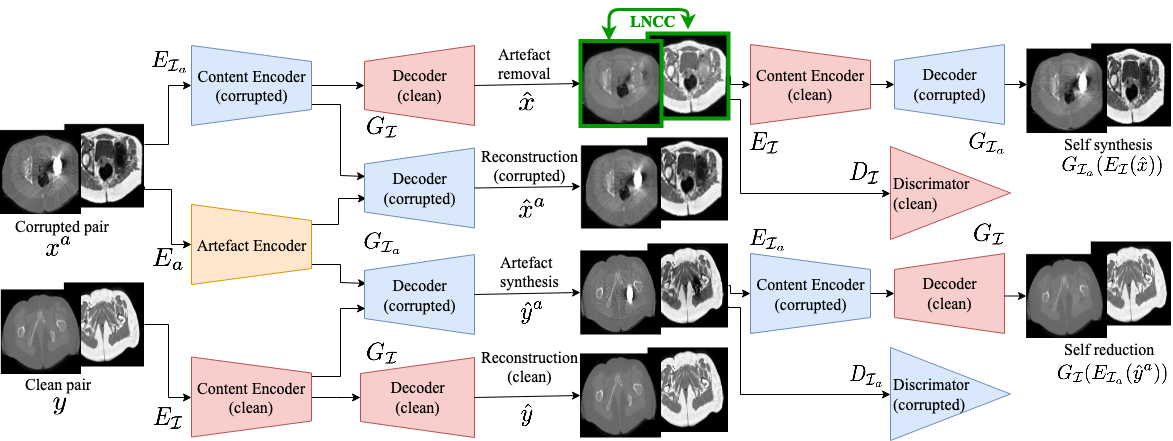}
  \caption{Schematic representation of MADN. We used a multichannel ADN architecture~\cite{Liao2019} and introduced a novel similarity loss term to adapt it to the multimodal scenario.}
  \label{fig:arch}
\end{center}
\end{figure*}
Our work builds upon the Artefact Disentanglement Network (ADN) recently proposed by Liao \etal~\cite{Liao2019} to perform unsupervised MAR on CT images. The ADN uses two sets of unpaired images, one including metal artefact corrupted CTs and one with clean non-corrupted images. The key idea is to use encoder-decoder networks coupled with adversarial training to learn a latent representation of the data where the artefact is disentangled from the anatomical content. This separation allows to reconstruct the corrupted images using only the latent content representation, therefore removing the artefact. It also allows to reconstruct the denoised images with the latent artefact representation and thus synthesising corrupted images. \\
Let $\mathcal{I}$ and $\mathcal{I}_a$ be the domains of clean and corrupted images respectively. The network architecture (Fig. \ref{fig:arch}) is composed as follows: Three encoders ($\mathrm{E}_{\mathcal{I}}: \mathcal{I} \rightarrow \mathcal{C}$, $\mathrm{E}_{\mathcal{I}_a}: \mathcal{I}_a \rightarrow \mathcal{C}$, $\mathrm{E}_{a}: \mathcal{I}_a \rightarrow \mathcal{A}$) map the input images to either the content $\mathcal{C}$ or the artefact $\mathcal{A}$ latent spaces; two decoders map the latent space back to the image domain and work as generators ($\mathrm{G}_{\mathcal{I}}: \mathcal{C} \rightarrow \mathcal{I}$, $\mathrm{G}_{\mathcal{I}_a}: \mathcal{C}\times\mathcal{A}\rightarrow \mathcal{I}_a$); finally two discriminators ($\mathrm{D}_{\mathcal{I}}$, $\mathrm{D}_{\mathcal{I}_a}$) define whether an input is real (i.e. coming from the real distribution of $\mathcal{I}$ or $\mathcal{I}_a$ respectively) or fake (i.e. synthetically generated by the decoders $\mathrm{G}_{\mathcal{I}}$ and $\mathrm{G}_{\mathcal{I}_a}$ respectively). We refer the reader to the original manuscript~\cite{Liao2019} for the layer-by-layer outline of the network. \\
Given a corrupted image $x^a \in \mathcal{I}_a$ and a clean image $y \in \mathcal{I}$, we can define their encoding as $c^a = \mathrm{E}_{\mathcal{I}_a}(x^a), a = \mathrm{E}_a(x^a)$ and $c = \mathrm{E}_{\mathcal{I}}(x)$. Indicating by $\hat{}$ the decoded images, from $x^a$ we obtain a reconstructed corrupted image $\hat{x}^a = G_{\mathcal{I}_a}(c^a, a)$ and its corrected version  $\hat{x} = G_{\mathcal{I}}(c^a)$. Similarly, $\hat{y}^a = G_{\mathcal{I}_a}(c, a)$ is the synthetically corrupted image from the clean input and $\hat{y} = G_{\mathcal{I}}(c)$ is the reconstructed clean image. 
To guarantee the expected outputs, the network is trained to minimise the following total loss function:
\begin{equation}
 \mathcal{L}_{tot} = \lambda_{adv}^{\mathcal{I}} \mathcal{L}_{adv}^{\mathcal{I}} + \lambda_{adv}^{\mathcal{I}_a} \mathcal{L}_{adv}^{\mathcal{I}_a} + \lambda_{rec}\mathcal{L}_{rec} + \lambda_{sr}\mathcal{L}_{sr} + \lambda_{art}\mathcal{L}_{art}   
\end{equation}
The first two terms are the traditional adversarial losses that promote a realistic generation of clean and corrupted images from $G_{\mathcal{I}}$ and $G_{\mathcal{I}_a}$:
\begin{equation}
\begin{split}
    \mathcal{L}_{adv}^{\mathcal{I}} & = \mathbb{E}_{\mathcal{I}}[\log D_{\mathcal{I}}(y)] + \mathbb{E}_{\mathcal{I}_a}[1 - \log D_{\mathcal{I}}(\hat{x})] \\
\mathcal{L}_{adv}^{\mathcal{I}_a} &= \mathbb{E}_{\mathcal{I}_a}[\log D_{\mathcal{I}_a}(x)] + \mathbb{E}_{\mathcal{I},\mathcal{I}_a}[1 - \log D_{\mathcal{I}_a}(\hat{y}^a)] 
\end{split}
\end{equation}
The reconstruction loss guarantees that same-branch encoding-decoding correctly reconstructs the input, thus ensuring the preservation of patient's anatomy:
\begin{equation}
    \mathcal{L}_{rec} = \mathbb{E}_{\mathcal{I},\mathcal{I}_a} [||\hat{x}^a - x^a||_1 + ||\hat{y} - y||_1 ]
\end{equation}
The self-reduction loss promotes cycle consistency within the cycle ``clean - corrupted - clean":
\begin{equation}
    \mathcal{L}_{sr} = \mathbb{E}_{\mathcal{I},\mathcal{I}_a} [||G_{\mathcal{I}}(E_{\mathcal{I}_a}(\hat{y}^a)) - y ||_1 ]
\end{equation}
Finally, the artefact consistency loss enforces that the artefact removed through the denoising path is the same added from the artefact-synthesis path, \textit{de facto} training $E_a$ to encode the artefact only:
\begin{equation}
    \mathcal{L}_{art} = \mathbb{E}_{\mathcal{I},\mathcal{I}_a} [||(x^a - \hat{x}) -(\hat{y}^a - y)||_1 ]
\end{equation}
Each loss term is weighted by the respective hyper-parameter $\lambda$. 

\subsection{Multimodal Artefact Disentanglement Network}
In this work, the ADN is extended to a multimodal case, using two-channel inputs $x^a$ and $y$, with CT image as first channel and respective registered MRI as second channel. The network learns to correct for the artefact on both modalities simultaneously, using multimodal information to encode the anatomical content of the images. To further enforce this sharing of information between the modalities, we introduce the use of a loss term to maximise the similarity of the artefact-corrected images. This is motivated by the idea that two different images of the same object appear less similar if corrupted by noise or artefacts, especially when the artefacts present with different patterns in the two images. Conversely, the two images should look more similar if artefact-free. 
By maximising the similarity between the output channels, we aim to improve artefact reduction for both modalities: firstly, the high-frequency and full-field-of-view nature of the artefact in the CT could be corrected through comparison with artefact-free MRI regions; secondly, the implant lack of signal in MRI could be compensated by the CT information, and better reconstruction should be achieved. 
We choose Locally Normalised Cross Correlation (LNCC) as a measure of similarity, as it is suitable for multimodal comparison and it can be efficiently incorporated onto a neural network framework thanks to its convolution formulation~\cite{Cachier2003}. The new similarity loss term is thus defined as:
\begin{equation}
    \mathcal{L}_{sim} = 1 - \mathbb{E}_{\mathcal{I}_a}[|LNCC(\hat{x}_{CT}, \hat{x}_{MRI})|]
\end{equation}
In addition, we also consider a self-synthesis consistency loss for the cycle ``corrupted - clean - corrupted", that constitutes a full cycle loss together with the self-reduction loss:
\begin{equation}
    \mathcal{L}_{cycle} =  \mathcal{L}_{sr} +  \mathbb{E}_{\mathcal{I}_a} [||G_{\mathcal{I}_a}(E_{\mathcal{I}}(\hat{x}), a) - x^a ||_1 ]
\end{equation}
In our experience, this helps obtain sharper output images, especially with small training set size. 
The final total loss for training the MADN architecture is thus 
\begin{multline}
 \mathcal{L}_{tot} = \lambda_{adv}^{\mathcal{I}} \mathcal{L}_{adv}^{\mathcal{I}} + \lambda_{adv}^{\mathcal{I}_a} \mathcal{L}_{adv}^{\mathcal{I}_a} + \lambda_{rec}\mathcal{L}_{rec} + \\
 \lambda_{cycle}\mathcal{L}_{cycle} + \lambda_{art}\mathcal{L}_{art} + \lambda_{sim}\mathcal{L}_{sim}
\end{multline}

\subsection{Experimental Setup}
\label{ssec:subhead}
Our dataset included 65 3D CT-T1 MARS MRI pairs from subjects with metal hip implants, and 63 CT-T1 MRI clean pairs, with no metal artefacts. Data was collected in clinical setting, with a variety of acquisition protocols, in compliance with the Helsinki Declaration. Each CT-MRI pair has been aligned with non-linear registration using a cubic b-spline free-form deformation algorithm~\cite{Modat2010}. Note that the CT images were initially corrected with the Refined Metal Artefact Reduction (RMAR)~\cite{treece2017} for more accurate transformation estimation. However, the uncorrected images were utilised in all the experiments subsequently described. 11 pairs of corrupted CT-MRI were associated with manual segmentation of four muscles - Gluteus Maximus (GMAX), Gluteus Medius (GMED), Gluteus Minimus (GMIN) and Tensor Fasciae Latae (TFL). These subjects were thus left out from training and used as test set.\\
To quantify the effect of the MAR on CT we computed the standard deviation of the intensities ($\sigma_{CT}$) within the muscle regions. The presence of metal artefact induces noise even further from the implant, causing fluctuations of the intensities from their true value, and therefore higher standard deviation. We performed this analysis before correction (No MAR) and after correction with: (RMAR CT) a conventional MAR algorithm~\cite{treece2017}; (ADN CT) correction using an ADN model trained on CT only; (Multichannel ADN) correction using an ADN two-channel model trained on CT and MRI; (MADN) our proposed correction based on two-channel ADN model with LNCC similarity loss. 
For the MRI, we performed a segmentation propagation experiment: each test MRI was registered to all others using an intensity-based free-form deformation algorithm~\cite{Modat2010}, their manual segmentation was propagated with the estimated transformation and compared with the manual ground truth using the Dice score. In addition to No MAR, Multichannel ADN and MADN, for this task we also trained an ADN model using MR only (ADN MR).
All ADN models were trained on 2D slices with ADAM optimiser and learning rate $=10^{-5}$. The loss weights were set to $\lambda_{adv}^{\mathcal{I}}=\lambda_{adv}^{\mathcal{I}_a} = 1.0 $, $\lambda_{cycle} = \lambda_{rs}=\lambda_{rec} = \lambda_{art} = 20.0$. For the proposed MADN, we set $\lambda_{sim} = 1.0$, and LNCC estimated through a Gaussian kernel with $\sigma=5$.
\begin{figure*}[t!]
\begin{center}
  \includegraphics[width=0.87\textwidth,height=4.5cm]{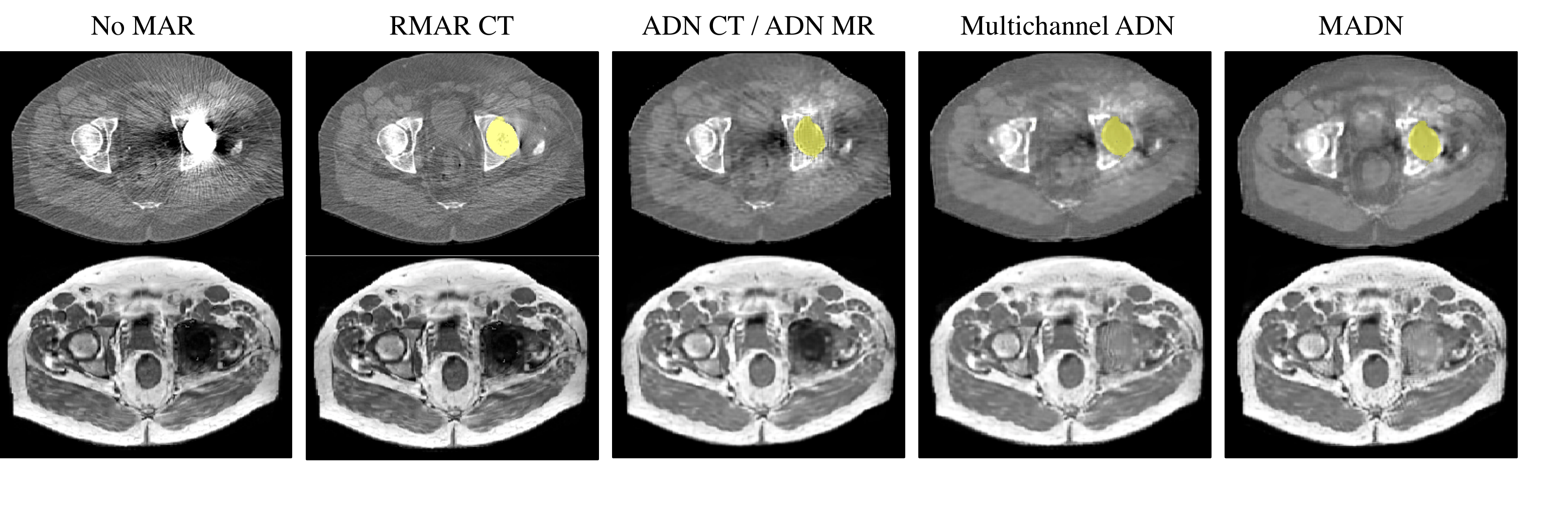}
  \caption{Visual comparison of MAR methods on CT and MRI. The implant is highlighted in yellow in all the corrected CT images. No overlay is applied to the MRI to better display the effect of the MAR.}
  \label{fig:qualitative}
\end{center}
\end{figure*}

\section{Results and discussion}
\label{sec:conc}

\begin{figure}[h!]
\subfigure[]{\includegraphics[width=8.8cm,height=4.6cm]{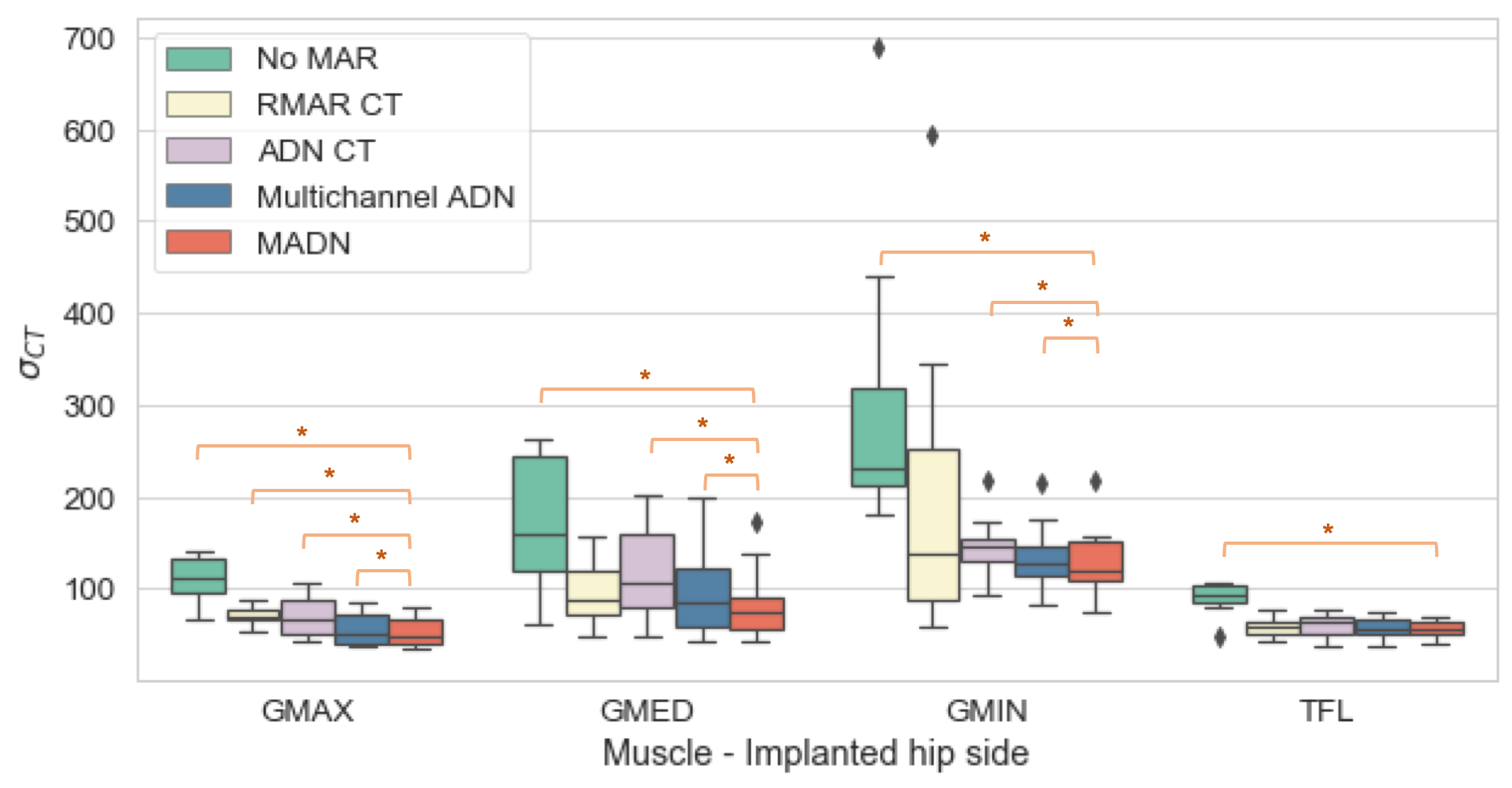}}
\subfigure[]{\includegraphics[width=8.8cm,height=4.6cm]{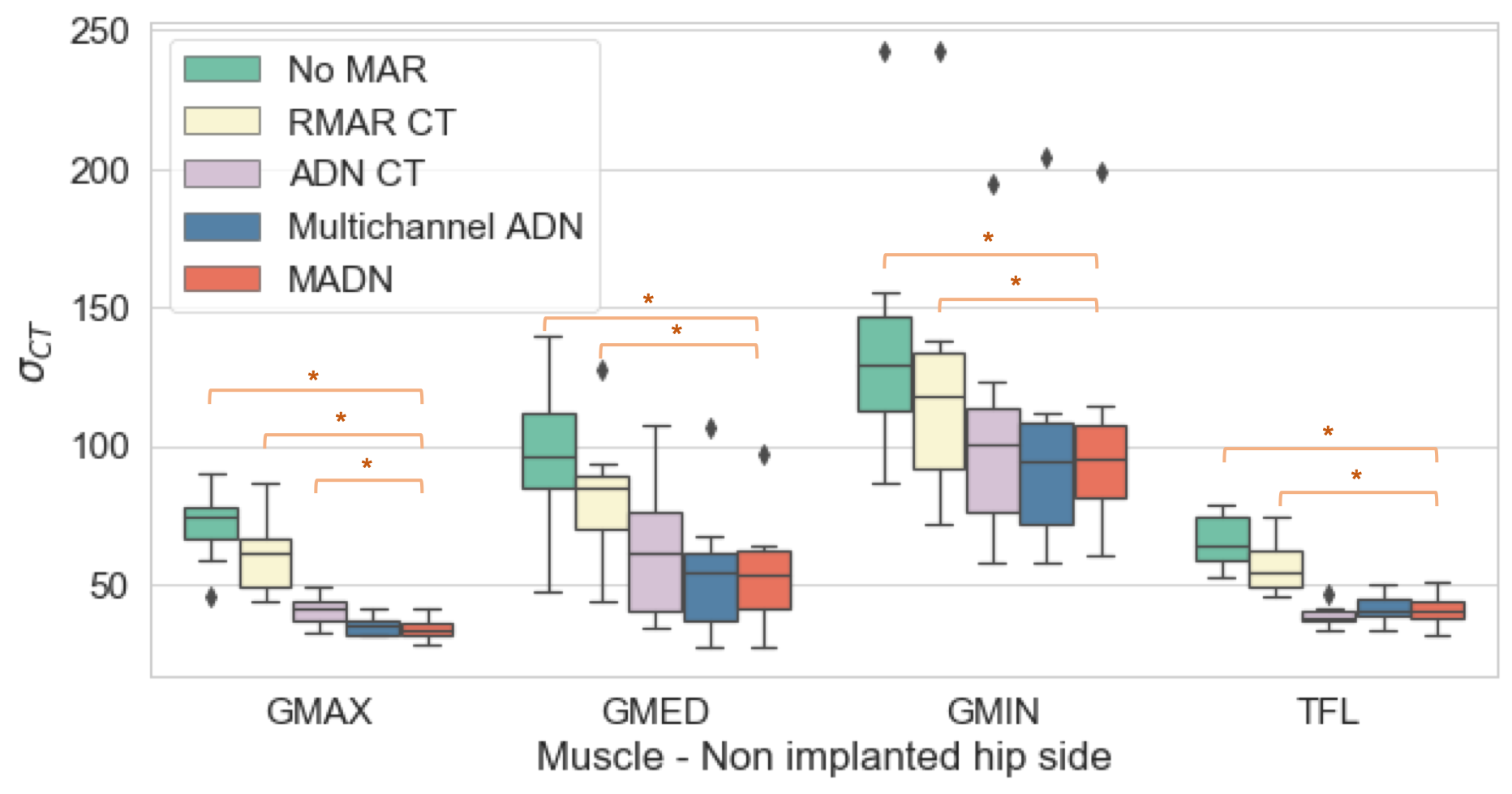}}
\captionsetup{belowskip=-2pt}
\caption{Standard deviation of CT intensity values within specific muscles. Cases significantly different from MADN are indicated by * (two-tailed paired t-test with $p<0.05$). (a) Implanted hip side. (b) Non implanted hip side.}
\label{fig:resCT}
\end{figure}

Figure \ref{fig:qualitative} presents a visual comparison of all the tested MAR methods. On the CT, our approach is the most effective in reducing the streaks artefacts throughout the full field of view, as the correction is also driven by non-corrupted MRI corresponding areas. This reduction is also demonstrated by the decrease in $\sigma_{CT}$ within the muscular tissue (Fig. \ref{fig:resCT}), on either the implanted and non-implanted hip sides.
\vfill

On the MRI, Fig. \ref{fig:qualitative} shows that training with MR images only (ADN MR) is not sufficient to learn an embedding of the artefact and therefore correct for it. The multichannel ADN and MADN approaches instead identify the corrupted area correctly and attempt to reconstruct the signal in it. However, the shape of the implant or the surrounding tissue is still not fully recovered. The quantitative experiments reported in Table \ref{tab:MR_to_MR_dice} and Fig. \ref{fig:mr_to_mr} show that MADN provides slightly better alignment for GMAX and TFL, but it performs worse on GMIN. It is however worth noticing that the manual segmentations were performed on the non-corrected MRI, where GMIN is the most affected by the artefact. This makes it challenging to determine whether such result is due to less accurate registration or unreliable ground truth. Further analysis is thus needed to better quantify the MAR impact on MRI. \\
Together with the lack of a clear ground truth, a few limitations characterise our study. First, the ADN architecture was not modified, keeping the same amount of parameters for either single- or multi-modality tests and thus not optimising the model capacity to the multimodal task. Moreover, different multimodal similarity measures could be implemented in place of LNCC (e.g. Cross Correlation or Normalised Mutual Information) to test their impact on the reconstruction. Finally, the test set is currently limited in size due to lack of ground truth for quantitative analysis. Hence, the generalisability of the proposed approach is to be more extensively validated.
Despite these limitations, the qualitative examples and quantitative results on the CT suggest that the use of the multimodal approach for MAR could be beneficial, as it combines different information to learn a better embedding of the anatomy and of the artefact. Future work will thus address the aforementioned limitations and focus on hyperparameters optimisation in order to improve the correction in MRI. 
\begin{table}[t!]
\begin{minipage}[b]{1.0\linewidth}
  \centering
\resizebox{8.5cm}{!}{
\begin{tabular}{l|c|c|c|c}
                 & GMAX                  & GMED                  & GMIN                  & TFL  \\ \hline
No MAR           & 0.76$\pm$0.15          & \textbf{0.54}$\pm$0.17 & \textbf{0.37}$\pm$0.18 & 0.41$\pm$0.27          \\
ADN MR           & 0.76$\pm$0.18          & 0.53$\pm$0.16  & 0.34$\pm$0.19  & 0.44$\pm$0.26          \\
Multich ADN & \textbf{0.77}$\pm$0.17 & 0.53$\pm$0.16          & 0.31$\pm$0.19          & 0.46$\pm$0.27          \\
MADN             & \textbf{0.77}$\pm$0.16 & \textbf{0.54}$\pm$0.15 & 0.32$\pm$0.19          & \textbf{0.50}$\pm$0.26 \\ \hline
\end{tabular}
}
\caption{Mean and standard deviation of Dice score for MR-to-MR inter-subject segmentation propagation task.}
\label{tab:MR_to_MR_dice}
\vspace{0.25cm}
\end{minipage}
\begin{minipage}[b]{1.0\linewidth}
    \centering
    \centerline{\includegraphics[width=8cm, height=4.5cm]{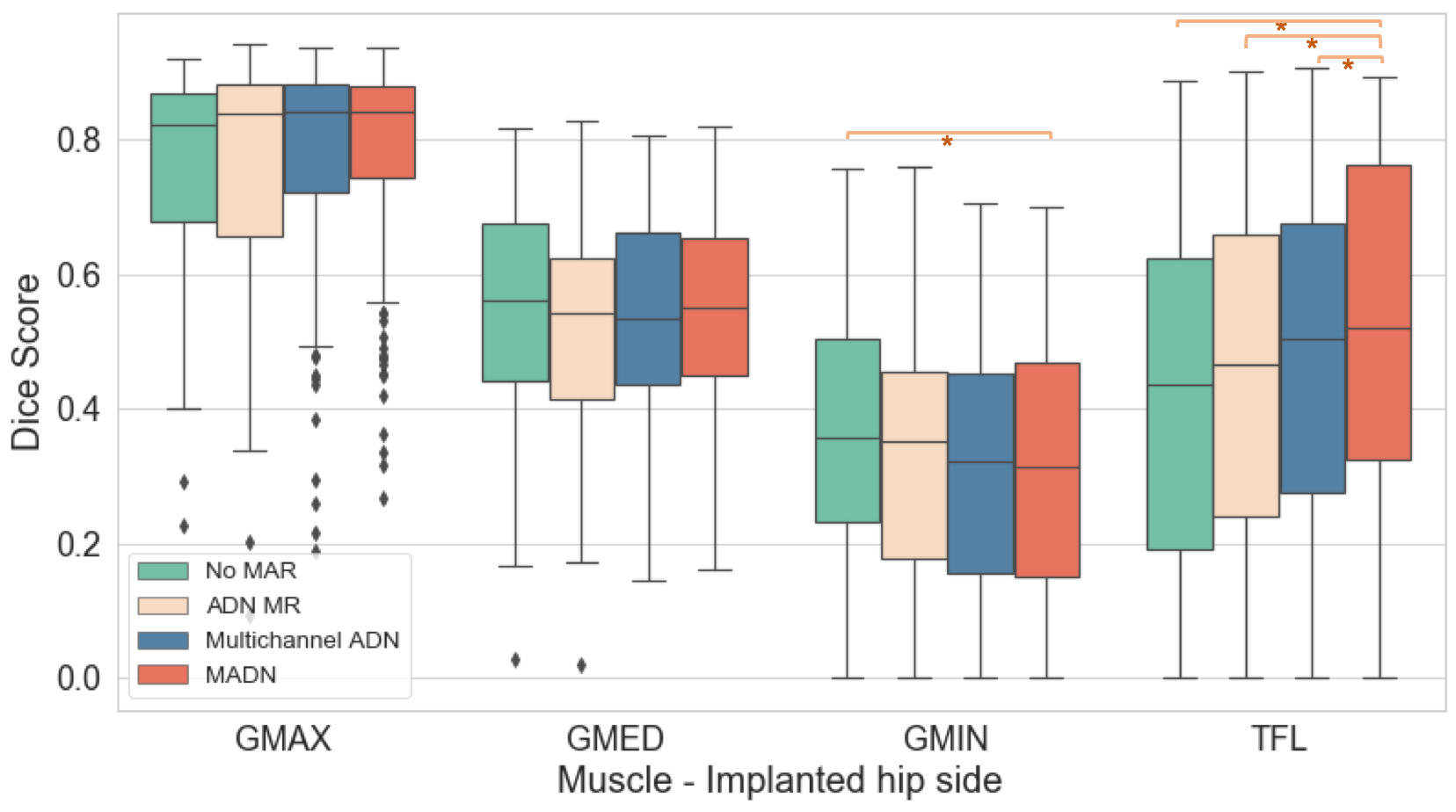}}
    \captionsetup{belowskip=0pt}
    \captionof{figure}{Dice score obtained from MR-to-MR inter-subject registration and segmentation propagation. Cases significantly different from MADN are indicated by * (two-tailed paired t-test with $p<0.05$)}
    \label{fig:mr_to_mr}
\end{minipage}
\end{table}\vfill
\paragraph*{Acknowledgements}
This work is supported by the EPSRC-funded UCL CDT in Medical Imaging [EP/L016 478/1], the RNOH NHS Trust, the Wellcome/EPSRC Centre for Medical Engineering [WT 203148/Z/16/Z, NS/A000049/1] and the NIHR Biomedical Research Centre based at GSTT NHS Trust and KCL.
\bibliographystyle{IEEEbib}
\balance
\bibliography{strings}

\end{document}